%% file: main.tex
\documentclass[letterpaper,10pt,conference]{ieeeconf}

\IEEEoverridecommandlockouts
\usepackage{amsmath}
\usepackage{amssymb}
\usepackage{algorithm}
\usepackage{algpseudocode}
\usepackage{booktabs}
\usepackage{cite}
\usepackage{graphicx}
\usepackage{microtype}
\usepackage{multirow}
\usepackage{times}
\usepackage{url}
\usepackage{xcolor}
\usepackage{xspace}
\usepackage{balance}
\usepackage{tikz}
\usetikzlibrary{arrows.meta,positioning,fit,calc}

\newcommand{\method}{\textsc{Redirect}\xspace}
\newcommand{\thetaInit}{\theta_{0}}
\newcommand{\datasetRetain}{\mathcal{D}_{r}}
\newcommand{\datasetDesignated}{\mathcal{D}_{d}}
\newcommand{\datasetEdit}{\mathcal{D}_{e}}

\title{\LARGE \bf
REDIRECT: A 1\% Fix for Bad Robot Habits
}

\author{Yu Zhang$^{1}$, Jiazhuo Li$^{2}$, Yancong Wei$^{1}$, Kangkang Dong$^{1}$,\\
Xiaojun Zhu$^{1}$, and Houde Liu$^{1,*}$%
\thanks{$^{1}$Tsinghua University.}%
\thanks{$^{2}$University of Michigan.}%
\thanks{$^{*}$Corresponding author: Houde Liu.}%
}

\begin{document}

\maketitle
\thispagestyle{empty}
\pagestyle{empty}

\begin{abstract}
Robots can acquire bad habits from a few defective moments in otherwise useful teleoperation. In mixed-quality robot data, normal and problematic demonstrations share most task behavior and differ only at a local action continuation. Full retraining is costly, while fine-tuning on clean data alone offers limited recovery under a small update budget. We ask whether the local difference can instead be fixed with only 1\% of the full-training sample-backward budget. Using only episode-level retained/problematic labels, \method localizes the branch, assigns a coherent retained continuation to the original observations around it, and anchors shared behavior, without frame-level annotations or additional interaction. Across three randomized ManiSkill tasks and three seeds, \method raises the mean success rate from 68.2\% to 90.3\%, recovering 88.8\% of the clean-retraining gap versus 22.1\% for matched-compute fine-tuning. On a PiPER arm, it recovers 86.7--92.9\% of the clean-only gap across cup insertion and towel folding. Local robot habits can therefore be repaired by spending the update budget on the behavioral difference rather than relearning shared behavior.
\end{abstract}

\input{sections/01_introduction}
\input{sections/02_related_work}
\input{sections/03_method}

\input{sections/04_experiments}
\input{sections/05_results}
\input{sections/06_discussion}
\input{sections/07_conclusion}

\bibliographystyle{IEEEtran}
\bibliography{references}

\end{document}

%% file: sections/01_introduction.tex
\section{Introduction}
\label{sec:introduction}

A robot may insert a cup correctly yet repeatedly tip it over during release. Such failures can arise from brief teleoperation errors even when most of the demonstrated behavior remains useful. Full retraining incurs substantial computational cost, whereas fine-tuning on clean data alone may offer limited correction under a small update budget. This motivates concentrating repair compute on the local difference.

Can this difference be repaired with only 1\% of the full-training sample-backward budget? We study ordinary and locally flawed demonstrations that execute the same task and share most behavior. A \emph{bad robot habit} is a repeatable action-chunk deviation within an otherwise task-consistent trajectory. The auditor provides only episode-level labels; retained episodes supply normal continuations.

In this setting, the local difference naturally induces a \emph{behavioral branch}: a problematic demonstration follows retained-compatible behavior until a local decision separates its action future and shifts later observations. Localizing this separation under unpaired trajectory variation isolates the repair from shared behavior.

\begin{figure*}[t]
    \centering
    \includegraphics[width=\textwidth]{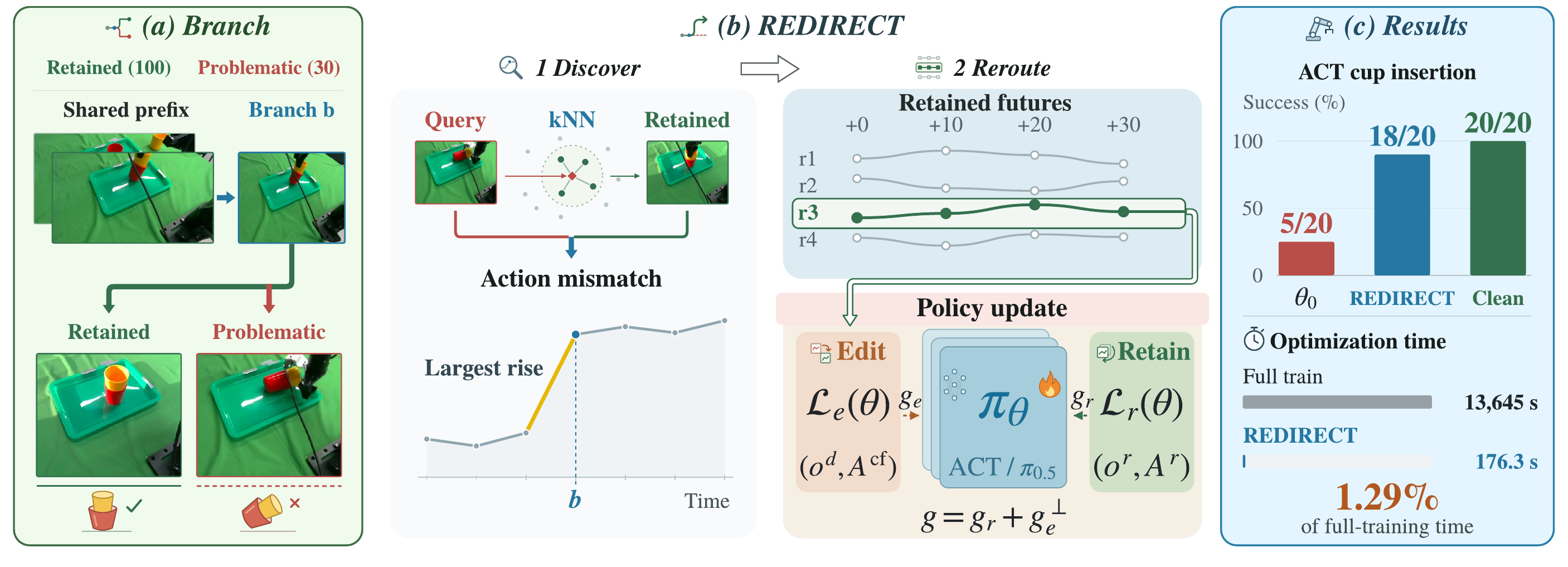}
    \caption{Overview of \method at the 1\% operating point. (a) Retained and problematic demonstrations share a prefix, then diverge at branch $b$ into retained and failed continuations. (b) Schematic: \method extracts motion-transition candidates, retrieves comparable retained futures using kinematics and RGB, localizes the largest rise in clean-calibrated future-action departure, selects one coherent retained continuation, and performs an edit/retain-anchored local update. (c) On ACT real-robot cup insertion, \method improves $\thetaInit$ from 5/20 to 18/20 successes, close to clean-only at 20/20, while requiring 176.3~s of optimization versus 13,645~s for full training (1.29\%).}
    \label{fig:method}
\end{figure*}

Fine-tuning on retained data does not directly provide corrective action targets at observations from problematic demonstrations. \method localizes the branch, pairs observations around it with a coherent retained continuation, and anchors each update on retained demonstrations (Fig.~\ref{fig:method}).

The comparison is clean-conditioned but unpaired: motion, state, and RGB descriptors identify context-compatible retained futures, while action-future departure determines the branch. Synthetic branch timestamps are reserved for evaluation rather than supplied to the repair procedure.

At 1\% of full-training sample-backward compute, \method raises the mean success rate in simulation from 68.2\% to 90.3\% and recovers 88.8\% of the clean gap, versus 22.1\% for matched-compute fine-tuning. Across ACT, $\pi_{0.5}$, cup insertion, and towel folding, it recovers 86.7--92.9\% of the clean-only gap with branch-local changes.

This paper makes three contributions:
\begin{itemize}
    \setlength{\itemsep}{-1pt}
    \setlength{\parsep}{0pt}
    \setlength{\topsep}{2pt}
    \item We formulate low-budget policy repair around behavioral branches in mixed-quality demonstrations.
    \item We introduce \method: episode membership localizes the separation, one trajectory-consistent normal continuation reroutes it, and retained supervision anchors shared behavior without reward, rollout, or teacher.
    \item Across ACT, $\pi_{0.5}$, simulation, and real robots, 1\% of full-training sample-backward compute recovers 88.8\% of the simulation gap with branch-local changes.
\end{itemize}

%% file: sections/02_related_work.tex
\section{Related Work}
\label{sec:related}

\paragraph{Interactive and post-training policy correction}

Behavior-cloning errors compound off distribution~\cite{ross2011dagger}. Interactive methods including DAgger and HG-DAgger~\cite{ross2011dagger,kelly2019hgdagger}, intervention-weighted regression~\cite{mandlekar2020iwr}, and ThriftyDAgger~\cite{hoque2022thriftydagger} improve policies with expert interventions; RaC scales recovery--correction segments~\cite{hu2025rac}. Residual RL adds reward-driven refinement~\cite{ankile2025resip}, while set-supervised diffusion learns from paired corrections~\cite{li2026setsupervised}. These approaches focus supervision on difficult states but require new rollouts, interventions, corrective labels, or rewards. ACT, Diffusion Policy, and the flow-matching-based $\pi_0/\pi_{0.5}$~\cite{zhao2023act,chi2023diffusion,lipman2023flow,black2024pi0,physicalintelligence2025pi05} are policy backbones rather than repair procedures. \method instead performs offline post-training repair from episode-level retained/problematic labels and assigns a fixed budget to their local action-future branch.

\paragraph{Learning from imperfect and mixed-quality demonstrations}

Confidence-weighted imitation learning~\cite{wu2019imperfect} and VILD~\cite{tangkaratt2020vild} address imperfect or latent-quality demonstrations. Demo-SCORE curates demonstrations using online experience~\cite{chen2025demoscore}, IntervenGen expands a small intervention set~\cite{hoque2024intervengen}, and S2I selects and refines demonstration segments through trajectory optimization and action relabeling~\cite{chen2025s2i}. Recent curation methods assess demonstration quality through state--action mutual information~\cite{hejna2025deminf} or estimated influence on closed-loop return~\cite{agia2025cupid}. \method instead retains both episode sets and repairs an already-trained policy by localizing the dominant departure and assigning a retained continuation under a fixed update budget.

\paragraph{Retrieval and temporal localization}

Behavior Retrieval queries unlabeled datasets for few-shot imitation~\cite{du2023behavior}, while VINN predicts actions from nearby visual demonstrations~\cite{pari2022vinn}. CompILE discovers reusable variable-length temporal units in demonstrations~\cite{kipf2019compile}. \method uses retrieval neither as the deployed controller nor as generic skill discovery: it matches clean-conditioned futures to localize a branch onset and selects a single-source, time-consistent continuation as the local editing target.

\paragraph{Retention-aware local adaptation}

LoRA reduces trainable parameters~\cite{hu2022lora}, GradNorm balances gradient magnitudes~\cite{chen2018gradnorm}, PCGrad projects conflicting objectives~\cite{yu2020pcgrad}, and GEM/A-GEM constrain updates with memory gradients~\cite{lopezpaz2017gem,chaudhry2019agem}. These tools address adaptation efficiency or interference, but do not determine where a robot policy should be edited or what executable behavior should replace a defect. \method couples branch localization, coherent continuation retrieval, retained replay, gradient balancing, and conflict projection to concentrate a fixed sample-backward budget while limiting drift on shared behavior.

%% file: sections/03_method.tex
\section{\method: A One-Percent Behavioral Fix}
\label{sec:method}

\subsection{Fixed-Budget Problem Formulation}

\textbf{Mixed-quality teleoperation structure.} Let $\datasetRetain=\{\tau_i^r\}$ contain ordinary demonstrations and $\datasetDesignated=\{\tau_i^d\}$ locally flawed demonstrations used to train $\pi_{\thetaInit}$. Both execute the same task and share most behavior. A trajectory in $\datasetDesignated$ follows ordinary behavior until a local decision departs from a continuation in $\datasetRetain$; downstream observations can then diverge. In this setting, the local departure naturally induces a behavioral branch. Only episode membership is given---no branch frame, subtask, corruption category, or interval. We localize and reroute the dominant branch while preserving shared behavior under a small budget. No clean policy is queried; retained demonstrations define the normal continuation.

For action horizon $H$, let $A_t=[a_t,\ldots,a_{t+H-1}]\in\mathbb{R}^{H\times d_a}$.

We measure repair cost by sample-backward (SB) evaluations. Our primary simulation operating point fixes the repair budget to 1\% of full training; \method itself is not tied to a particular update count. The objective is to maximize closed-loop recovery under this fixed budget while preserving useful behavior.

\subsection{Clean-Conditioned Branch Localization}

A bad habit is characterized by a change point. It begins when a problematic continuation first departs from a retained-compatible future. Dense frames can let weak noise dominate localization, so we extract ordered motion-transition candidates
$\mathcal{C}_i=\{c_{i,1}<\cdots<c_{i,m_i}\}$ from recorded full actions. Candidates capture motion starts (mean absolute action change $>10^{-4}$) and direction reversals (cosine $<-0.5$, followed by three moving transitions); no task stage, gripper-event name, corruption label, or smoothing is used.

From retained data only, we robust-scale state and lag-5 velocity and form
\begin{equation}
 \psi_t=\left[\hat s_t,\widehat{s_t-s_{t-5}}\right].
 \label{eq:state-feature}
\end{equation}
Each candidate also has a synchronized front-camera descriptor $\phi_{\mathrm{rgb}}(o)$, obtained by deterministic per-channel-normalized RGB pooling.  To avoid matching a kinematically nearby but visually incompatible future, branch-reference and continuation matching use the same two-stage retrieval,
\begin{equation}
\begin{aligned}
 D^{\mathrm{kin}}_{cj}&=\|\psi^d_c-\psi^r_j\|_2^2,\\
 \mathcal Q_c&=\operatorname{Top128}_jD^{\mathrm{kin}}_{cj},\\
 D^{\mathrm{rgb}}_{cj}&=\|\phi_{\mathrm{rgb}}(o^d_c)-\phi_{\mathrm{rgb}}(o^r_j)\|_1,\\
 \mathcal N^{\mathrm{ctx}}_c&=\operatorname{TopK}_{j\in\mathcal Q_c}D^{\mathrm{rgb}}_{cj},\qquad K=4.
\end{aligned}
 \label{eq:scene-rerank}
\end{equation}
Neighbors come from distinct retained episodes. Simulation and physical $\pi_{0.5}$ use both stages; archived physical ACT lacks synchronized RGB and uses the four nearest kinematic candidates. Thus vision selects the clean reference without privileged object state. For medoid selection, $P(A)_{h,q}=(A_{h,q}-A_{0,q})/r_q$ on sampled horizon offsets $\mathcal H$, where $r_q$ is the retained action range. We select the medoid
\begin{equation}
 j_c^m=\arg\min_{j\in\mathcal{N}^{\mathrm{ctx}}_c}\frac{1}{K-1}
 \sum_{k\in\mathcal{N}^{\mathrm{ctx}}_c\setminus\{j\}}
 \left\|P(A_j^r)-P(A_k^r)\right\|_1
 \label{eq:frame-medoid}
\end{equation}
and measure departure in native action coordinates with the same retained scales,
\begin{equation}
 d_c=\frac{1}{|\mathcal H|d_a}\sum_{h\in\mathcal H}\sum_{q=1}^{d_a}
 \frac{|A^d_{c,h,q}-A^r_{j_c^m,h,q}|}{r_q}.
 \label{eq:discrepancy}
\end{equation}
The reference is chosen by relative-motion similarity, while $d_c$ retains absolute arm and gripper offsets. RGB affects reference matching through Eq.~\eqref{eq:scene-rerank}, while onset evidence remains a full-action departure; the original image observation also conditions every loss.

Clean leave-one-episode-out calibration of $d_c$ gives median $\mu_c$ and robust scale $\sigma_c=\max(1.4826\,\operatorname{MAD}_c,\sigma_{\min})$. Here, $\operatorname{MAD}_c$ is the median absolute deviation of clean scores from $\mu_c$; 1.4826 gives Gaussian-consistent scaling, and $\sigma_{\min}=0.005$ prevents near-zero denominators. Set $z_c=(d_c-\mu_c)/\sigma_c$ and define the onset jump along the ordered candidates
\begin{equation}
\begin{aligned}
 q_{i,m}&=z_{c_{i,m}}-z_{c_{i,m-1}}, \qquad q_{i,1}=z_{c_{i,1}},\\
 m_i^\star&=\min\arg\max_{1\leq m\leq m_i}q_{i,m},\\
 b_i&=c_{i,m_i^\star}.
\end{aligned}
    \label{eq:branch}
\end{equation}
Equation~\eqref{eq:branch} selects the observation-time edit anchor with the largest rise in future-action departure. The minimum resolves ties earlier; with several deviations, the one-pass formulation targets the dominant branch. Observations in $[b_i-w^-,b_i+w^+]$ enter the edit set. Synthetic oracle frames are used only for post-hoc auditing.

\subsection{Trajectory-Consistent Future Replacement}

Once the branch is localized, we construct a normal continuation at the problematic observation. Framewise matching can splice incompatible actions, while averaging multimodal futures can produce a continuation absent from the demonstrations. \method treats one retained trajectory as a single, coherent continuation whose identity and relative time remain fixed across the edit window. At $b_i$, Eq.~\eqref{eq:scene-rerank} yields $\mathcal N_i=\mathcal N^{\mathrm{ctx}}_{b_i}$. No object/goal state, external perception model, reward, or teacher supplies $\phi_{\mathrm{rgb}}$.

Let $\widetilde A_j=P(A_j^r)$. The representative source minimizes mean pairwise relative-motion distance,
\begin{equation}
 j_i^\star=\arg\min_{j\in\mathcal{N}_i}
 \frac{1}{K-1}\sum_{k\ne j}
 \left\|\widetilde A_j-\widetilde A_k\right\|_1.
 \label{eq:source-selection}
\end{equation}
The medoid is the most representative recorded normal continuation compatible with the branch context, not an external teacher.
Writing $j_i^\star$ for its frame index in the selected $T_i^r$-frame episode, temporal alignment maps designated offset $\Delta$ to
\begin{equation}
 \iota_i(\Delta)=\operatorname{clip}(j_i^\star+\Delta,0,T_i^r-1).
 \label{eq:alignment}
\end{equation}
The target directly copies the recorded arm and gripper commands,
\begin{equation}
 A^{cf}_{i,\Delta,h}=a^r_{\min(\iota_i(\Delta)+h,T_i^r-1)},
 \quad 0\leq h<H.
 \label{eq:counterfactual}
\end{equation}
Targets retain native recorded values with blend 1; policy-specific normalization follows target construction. Trajectory consistency means fixed source identity and relative time, with identical overlapping target actions wherever the source mapping advances one frame. Edit windows are clipped to the designated episode; Eq.~\eqref{eq:counterfactual} repeats the final source action at the boundary. The edit set $\datasetEdit=\{(o^d_{b_i+\Delta},A^{cf}_{i,\Delta})\}$ keeps the designated observation unchanged and replaces its local action future. Direct action replacement assumes comparable branch contexts under a shared task, controller, and action interface.

\subsection{Retain-Anchored Local Optimization}

Most behavior is shared and should remain unchanged: the edit objective rewrites the problematic continuation, while the retain objective anchors shared behavior,
\begin{align}
    \mathcal{L}_{e}(\theta)&=\mathbb{E}_{(o,A^{cf})\sim\datasetEdit}
    \ell_e\big(\pi_\theta(o),A^{cf}\big), \label{eq:edit-loss} \\
    \mathcal{L}_{r}(\theta)&=\mathbb{E}_{(o,A)\sim\datasetRetain}
    \ell_r\big(\pi_\theta(o),A\big). \label{eq:retain-loss}
\end{align}
Let $g_e=\nabla_\theta\mathcal{L}_e$ and $g_r=\nabla_\theta\mathcal{L}_r$.  Because the small edit set can produce gradients on a very different scale, we first set its norm relative to the retain gradient,
\begin{equation}
    \widetilde g_e=\rho\frac{\|g_r\|_2}{\|g_e\|_2+\epsilon}g_e.
    \label{eq:norm-balance}
\end{equation}
We then remove any component of the scaled edit gradient that opposes the retain objective:
\begin{equation}
    g_e^{\perp}=\widetilde g_e-
    \frac{\min(\langle\widetilde g_e,g_r\rangle,0)}{\|g_r\|_2^2}g_r,
    \qquad g=g_r+g_e^{\perp}.
    \label{eq:normpcgrad}
\end{equation}
This update is designed to limit interference with shared behavior. Here $\rho$ controls edit strength. Equation~\eqref{eq:normpcgrad} orthogonalizes conflict in Euclidean pre-optimizer space, but AdamW preconditioning and finite steps preclude a retention guarantee. We audit the complete step with
\begin{equation}
 C_{\mathrm{opt}}=\frac{1}{T}\sum_{t=1}^{T}
 \mathbf{1}\!\left[g_{r,t}^{\top}\Delta\theta_t>0\right],
 \label{eq:optimizer-conflict}
\end{equation}
where a positive dot product predicts a first-order increase of the current retain minibatch loss.  We ablate norm balancing and projection separately and jointly.

ACT uses full-chunk $L_1$ for $\ell_e$ and its native objective for $\ell_r$; $\pi_{0.5}$ uses flow matching for both and LoRA on the visual/observation-prefix context path.

\input{sections/algorithm_redirect}

%% file: sections/algorithm_redirect.tex
\begingroup
\setlength{\intextsep}{2pt}
\begin{algorithm}[H]
\footnotesize
\caption{\method: Fixed-Budget Behavioral Repair}
\label{alg:redirect}
\algrenewcommand\algorithmicindent{1em}
\algtext*{EndFor}
\algtext*{EndIf}
\begin{algorithmic}[1]
\Statex \textbf{Input:} $\theta_0$; retained $\mathcal D_r$; problematic $\mathcal D_d$
\Statex \textbf{Parameters:} $H,w^-,w^+,K,\rho,T$

\State $\theta\gets\theta_0$; $\mathcal D_e\gets\varnothing$
\State Index $\mathcal D_r$; calibrate $(\mu_c,\sigma_c)$ on $\mathcal D_r$ only
\For{$\tau_i^d\in\mathcal D_d$} \quad\textit{\# Discover}
  \State Extract motion-transition candidates $\mathcal C_i$
  \State Retrieve context-matched $\mathcal N_c^{\mathrm{ctx}}$, $c\in\mathcal C_i$ \hfill{\footnotesize\eqref{eq:scene-rerank}}
  \State Compute $d_c$; set $z_c\gets(d_c-\mu_c)/\sigma_c$ \hfill{\footnotesize\eqref{eq:frame-medoid}--\eqref{eq:discrepancy}}
  \State Select $b_i$ at the largest rise in $z_c$ \hfill{\footnotesize\eqref{eq:branch}}

  \State \textbf{Reroute:} $j_i^\star\gets\operatorname{medoid}(\mathcal N_{b_i}^{\mathrm{ctx}})$ \hfill{\footnotesize\eqref{eq:source-selection}}
  \For{valid $\Delta\in[-w^-,w^+]$}
    \State Copy $A^{cf}_{i,\Delta}$ from $j_i^\star$ at offset $\Delta$ \hfill{\footnotesize\eqref{eq:alignment}--\eqref{eq:counterfactual}}
    \State $\mathcal D_e\gets\mathcal D_e\cup\{(o^d_{b_i+\Delta},A^{cf}_{i,\Delta})\}$
  \EndFor
\EndFor

\For{$t=1,\ldots,T$} \quad\textit{\# Preserve}
  \State Sample $\mathcal B_e\sim\mathcal D_e$, $\mathcal B_r\sim\mathcal D_r$
  \State Compute $g_e=\nabla_\theta\mathcal L_e$, $g_r=\nabla_\theta\mathcal L_r$ \hfill{\footnotesize\eqref{eq:edit-loss}--\eqref{eq:retain-loss}}
  \State $\widetilde g_e\gets\rho\,\frac{\lVert g_r\rVert_2}{\lVert g_e\rVert_2+\epsilon}\,g_e$ \hfill{\footnotesize\eqref{eq:norm-balance}}
  \vspace{2pt}
  \If{$\langle\widetilde g_e,g_r\rangle<0$}
    \State Project $\widetilde g_e$ orthogonal to $g_r$ \hfill{\footnotesize\eqref{eq:normpcgrad}}
  \EndIf
  \State $\theta\gets\operatorname{AdamW}(\theta,g_r+\widetilde g_e)$
\EndFor
\State \Return $\pi_\theta,\ \{(b_i,j_i^\star)\}_i,\ T(|\mathcal B_e|+|\mathcal B_r|)$ SB
\end{algorithmic}
\end{algorithm}
\endgroup

%% file: sections/04_experiments.tex
\section{Experimental Setup}
\label{sec:experiments}

We evaluate repair performance, computational cost, and behavioral preservation across three simulation tasks and two real-robot tasks using ACT and $\pi_{0.5}$.

\subsection{Tasks and Demonstration Data}

Each setting starts from a policy $\thetaInit$ trained on 100 clean and 30 designated problematic demonstrations. Repair receives episode membership without temporal annotations.

\begin{figure*}[t]
    \centering
    \includegraphics[width=0.90\textwidth]{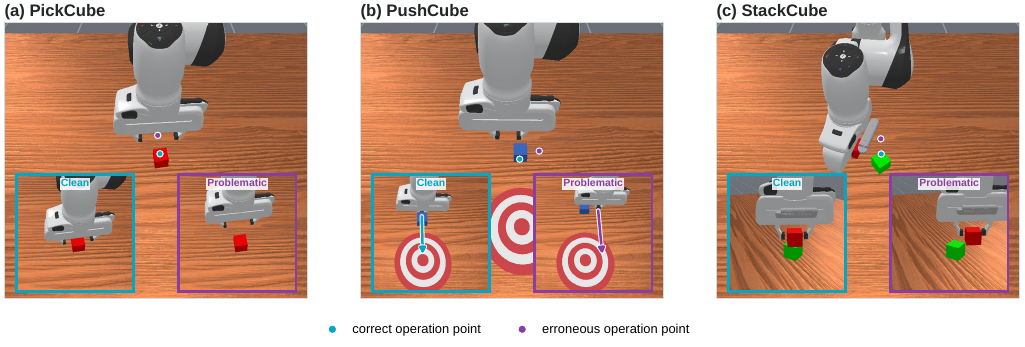}
    \caption{Randomized ManiSkill tasks with paired clean and problematic local behaviors. Correct (cyan) and erroneous (purple) operation points are marked in each scene; the lower-left and lower-right insets show matched clean and problematic intermediate states. PushCube arrows connect the current gripper center to the aligned or offset endpoint, while StackCube uses a side view to expose placement alignment.}
    \label{fig:sim_setup}
\end{figure*}

\textbf{Simulation.} We use ManiSkill PickCube, PushCube, and StackCube~\cite{gu2023maniskill2}, with local grasp, push, and placement offsets of 50/60/50~mm, respectively. Clean and designated data share the controller and reset distribution; 30 paired clean twins per task are reserved for auditing. Scenes are randomized (Fig.~\ref{fig:sim_setup}): PickCube independently samples cube and goal $xy$ within $\pm5$~cm; PushCube samples cube $xy$ within $\pm10$~cm and fixes the goal relative to the cube. Pick also varies cube yaw and goal height up to 15~cm. Stack varies both cube positions by $\pm3$~cm and randomizes both yaws. Joint-reset noise is 0.01/0.02/0.003~rad for Pick/Push/Stack. A separate heterogeneous StackCube set mixes 10 grasp-path, 10 transport-path, and 10 placement-pose deviations without defect-family labels.

\textbf{Real robots.} A 6-DoF PiPER arm with a parallel gripper and front/wrist RealSense cameras performs cup insertion and towel folding; state and action contain six joints and the gripper. Cup insertion places a yellow cup inside a red cup. The designated demonstrations share the approach and insertion but introduce an unstable release causing shaking, collision, or tip-over. Towel folding uses a left-over-right fold; its designated batch, Theta30, introduces a placement/release deviation that produces corner misalignment. ACT and $\pi_{0.5}$ are evaluated on cup insertion, and $\pi_{0.5}$ on towel folding.

\subsection{Training, Baselines, and Repair Budgets}

Full-training budgets are listed in Table~\ref{tab:wallclock}. Simulation training uses no oversampling; clean-only and mixed-data towel policies start from the same base model. All post-training methods use AdamW with the losses and backend adapters in Section~\ref{sec:method}. Within each backend, FT and \method start from the same mixed-data checkpoint.

Simulation repair uses $H=50$, window $[b-5,b+5]$, $\rho=0.32$, and learning rate $3\times10^{-5}$. All simulation post-training methods use 250 updates and 4,000 sample-backward (SB) evaluations, or 1\% of full training. \method and suppression use branch/retain batches of 8/8; FT-CM uses batch 16. Physical tasks use $[b-5,b+20]$: ACT uses learning rate $3\times10^{-5}$ and edit/retain batches 8/8; $\pi_{0.5}$ uses $2.5\times10^{-6}$ and batches 16/16. Physical update counts and budgets are listed in Table~\ref{tab:realrobot}.

FT-CM (Clean FT in Table~\ref{tab:realrobot}) uniformly samples retained observation--action pairs under the native loss, with the same SB budget as \method. Suppression combines retained replay with a reversed original branch-label loss in the same edit windows, without replacement targets. Physical ACT suppression uses three batch-8 terms, totaling 4,800 SB versus 3,200 for \method and FT. We additionally evaluate $\pi_{0.5}$ cup FT at four times the matched budget.

\subsection{Evaluation Protocol and Metrics}

Simulation configurations are frozen on seed-0 Dev50 (resets 9000--9049). Independent100 uses disjoint, unseen in-distribution resets 11000--11099, a 250-step limit, and identical resets across methods and three training seeds. Ablations hold the corresponding checkpoint, optimizer, hyperparameters, budget, and evaluation set fixed; preservation variants also share branches and targets. The heterogeneous StackCube experiment reuses the frozen configuration.

Each real checkpoint receives 20 trials at 30~Hz. Cup trials last 25~s, with $\pi_{0.5}$/ACT executing 50/25 actions per query. Cup success requires stable insertion of the yellow cup into the red cup after gripper release. Towel trials last 60~s with 50 actions per query; all checkpoints use the same ordered placements, varying towel center by 2--3~cm and yaw by $\pm5^\circ$. Terminal outcomes are upper/lower-corner miss, complete failure, or success; success requires a stable fold with both corners aligned. We report Wilson 95\% intervals, Fisher tests for cup trials, and exact McNemar tests for paired towel trials.

Alongside success $J$, we report point-estimate recovery of the gap between $\thetaInit$ and the clean-only reference:
\begin{equation}
 \operatorname{GapRec}(M)=
 \frac{J_M-J_{\thetaInit}}{J_{clean}-J_{\thetaInit}}\times100\%,
 \label{eq:gaprec}
\end{equation}

For binary outcomes $y_0(s)$ and $y_M(s)$ on the same simulation reset $s$, we measure the fraction of initial failures repaired and initial successes lost:
\begin{align}
 R_{\mathrm{repair}}(M)&=\frac{\sum_s(1-y_0(s))y_M(s)}{\sum_s(1-y_0(s))},\\
 R_{\mathrm{regress}}(M)&=\frac{\sum_s y_0(s)(1-y_M(s))}{\sum_s y_0(s)}.
 \label{eq:pairedrates}
\end{align}
Both rates are macro-averaged over task--seed pairs.

\subsection{Localization and Preservation Diagnostics}

Synthetic branch timestamps are reserved for localization audits. We replay nine simulation runs to check the recorded updates and evaluate optimizer-step conflict using Eq.~\eqref{eq:optimizer-conflict}.

For ACT, let $b$ denote the branch frame, $N$ the episode length, and $t_{\mathrm{rel}}$ the release frame, clamped to $[b+20,N-1]$ or set to $N-1$ if unavailable. Zero-indexed frames are partitioned into PRE ($0\leq t<b-5$), BRANCH ($b-5\leq t\leq b+20$), POST ($b+20<t\leq t_{\mathrm{rel}}$), and TAIL ($t_{\mathrm{rel}}<t<N$). We sample up to 26 evenly spaced observations per region per episode and measure original-label reconstruction $L_1$ over the first 25 predicted actions. Regional losses are averaged within each episode and then across 30 episodes, and normalized as:
\begin{equation}
    S_M(r)=\frac{L_M(r)-L_{\thetaInit}(r)}
    {L_{clean}(r)-L_{\thetaInit}(r)}.
    \label{eq:shift}
\end{equation}
Here, 0 and 1 denote the starting and clean-only label-loss levels. The separate ACT deployment audit uses the first 25 actions with retained $Q_{99}-Q_1$ scaling.

For $\pi_{0.5}$ diagnostics, each policy generates four $50\times7$ action chunks at each of 30 branch-center observations. Let $d$ denote MAE over all 350 entries, without per-dimension scaling or weighting. Problematic-pattern assignment is the fraction of these 120 predictions strictly closer to the original branch demonstration chunk than to a clean reference averaging future chunks from the eight nearest retained frames in 7-D state Euclidean distance, without RGB reranking.

Clean-directed progress instead uses predictions from the clean-only policy $C$ as reference. Let $D(M,C)$ average $d$ over corresponding observation--sample pairs. For \method $R$, progress is $100[1-D(R,C)/D(\thetaInit,C)]\%$, computed from mean distances rather than individual ratios. Figure~\ref{fig:mechanism} reports 95\% episode-bootstrap intervals, using 20,000 resamples for progress. Clean-only predictions and release timestamps are used only for diagnostics, not for localization, target construction, or repair updates.

%% file: sections/05_results.tex
\begin{figure*}[t]
    \centering
    \includegraphics[width=0.96\textwidth]{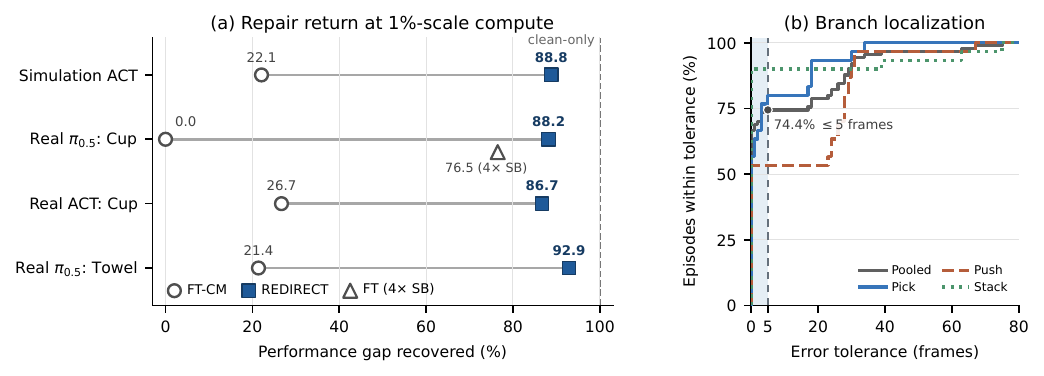}
    \caption{Automatic branch localization and repair return. At matched sample-backward budgets, \method converts a 1\%-scale update into substantially more clean-gap recovery than ordinary fine-tuning across simulation and real-robot settings; open circles denote matched clean FT, filled squares denote \method, and the triangle denotes four-times-SB $\pi_{0.5}$ fine-tuning. Panel (b) shows the empirical cumulative distribution of absolute localization error over 90 designated simulation episodes; the shaded band marks the $\pm5$-frame tolerance, and synthetic branch timestamps are used only for audit.}
    \label{fig:localization_compute}
\end{figure*}

\section{Results}
\label{sec:results}

\subsection{Simulation Repair Performance}

At the same 4,000-SB budget, \method raises mean success from 68.2\% to 90.3\%, recovering 88.8\% of the clean-only gap; FT-CM recovers 22.1\% (Table~\ref{tab:simulation}; Fig.~\ref{fig:localization_compute}(a)). \method performs best among post-training methods on all three tasks, with a 5.0\% regression rate (Table~\ref{tab:target_ablations}).

The improvement also holds when different episodes contain different defect types. On heterogeneous StackCube, \method reaches 81.0\% success and recovers 68.6\% of the clean-only gap without defect-family labels or retuning (Table~\ref{tab:heterogeneous}).

\begin{table}[t]
    \caption{One-percent repair on Independent100. Success (\%), mean $\pm$ SD over three training seeds. All post-training methods use the same 1\%-SB budget; bold denotes the best.}
    \label{tab:simulation}
    \centering
    \scriptsize
    \setlength{\tabcolsep}{1.7pt}
    \begin{tabular}{lrr|rrr}
        \toprule
        & \multicolumn{2}{c|}{Full training} & \multicolumn{3}{c}{1\% SB budget} \\
        \cmidrule(lr){2-3}\cmidrule(lr){4-6}
        Task & Clean & $\thetaInit$ & FT-CM & Supp. & \method \\
        \midrule
        PickCube  & $93.0{\pm}1.0$ & $65.7{\pm}2.9$ & $74.7{\pm}4.0$ & $79.3{\pm}7.5$ & $\mathbf{92.0{\pm}2.6}$ \\
        PushCube  & $96.0{\pm}4.0$ & $74.3{\pm}3.1$ & $71.7{\pm}11.0$ & $61.7{\pm}11.9$ & $\mathbf{92.7{\pm}1.2}$ \\
        StackCube & $90.3{\pm}2.3$ & $64.7{\pm}2.5$ & $74.7{\pm}7.2$ & $69.7{\pm}14.6$ & $\mathbf{86.3{\pm}4.7}$ \\
        \midrule
        Mean      & $93.1{\pm}1.3$ & $68.2{\pm}1.4$ & $73.7{\pm}4.1$ & $70.2{\pm}5.0$ & $\mathbf{90.3{\pm}2.6}$ \\
        \bottomrule
    \end{tabular}
\end{table}

\suppressfloats[t]
\begin{table}[t]
    \caption{Branch/target construction audit on Independent100 (macro \%); budget and optimizer are fixed.}
    \label{tab:target_ablations}
    \centering
    \scriptsize
    \setlength{\tabcolsep}{7pt}
    \begin{tabular}{lrrr}
        \toprule
        Variant & Success$\uparrow$ & RR$\uparrow$ & RG$\downarrow$ \\
        \midrule
        Random Window & 81.3 & 59.6 & 9.7 \\
        Local Clean FT & 77.8 & 44.0 & 7.7 \\
        K-neighbor average & 87.7 & 78.2 & 7.9 \\
        Per-chunk reselection & 82.0 & 71.3 & 12.9 \\
        Per-step reselection & 84.4 & 70.6 & 9.0 \\
        Shuffled relative time & 86.3 & 73.6 & 7.4 \\
        Full \method & \textbf{90.3} & \textbf{81.2} & \textbf{5.0} \\
        \bottomrule
    \end{tabular}
\end{table}

\begin{table}[t]
    \caption{Why the 1\% fix stays local: preservation-factor ablation on Independent100 (macro \%). All variants use 250 updates and 4,000 SB.}
    \label{tab:core_ablations}
    \centering
    \scriptsize
    \setlength{\tabcolsep}{1.8pt}
    \begin{tabular}{lccc|rrr}
        \toprule
        Variant & Ret. & Edit scale & Proj. & Succ.$\uparrow$ & RR$\uparrow$ & RG$\downarrow$ \\
        \midrule
        Starting $\thetaInit$ & -- & -- & -- & 68.2 & -- & -- \\
        \midrule
        Edit only & N & Raw & N & 39.0 & 33.4 & 58.6 \\
        Edit $+$ retain & Y & Raw & N & 86.0 & 75.5 & 8.5 \\
        Fixed $\lambda$ & Y & 0.32 & N & 85.3 & 78.8 & 11.4 \\
        Norm-balanced & Y & Norm & N & 88.0 & 76.9 & 6.7 \\
        Projected, no norm & Y & Raw & Y & 84.7 & 74.6 & 10.0 \\
        \midrule
        Full \method & Y & Norm & Y & \textbf{90.3} & \textbf{81.2} & \textbf{5.0} \\
        \bottomrule
    \end{tabular}
\end{table}

\begin{table}[t]
    \caption{Real-robot results. SB is sample-backward count; C/T denote cup/towel; intervals are Wilson 95\% CI; GapRec uses Eq.~\ref{eq:gaprec}.}
    \label{tab:realrobot}
    \centering
    \scriptsize
    \setlength{\tabcolsep}{0.9pt}
    \begin{tabular}{@{}llrrrrr@{}}
        \toprule
        Policy & Method & Upd. & SB & Succ. & Wilson CI & GapRec \\
        \midrule
        \multirow{5}{*}{$\pi_{0.5}$/C}
          & $\thetaInit$ & full & -- & 2/20 & [2.8,30.1] & 0.0 \\
          & Clean FT & 250 & 8000 & 2/20 & [2.8,30.1] & 0.0 \\
          & Clean FT & 1000 & 32000 & 15/20 & [53.1,88.8] & 76.5 \\
          & \method & 250 & 8000 & \textbf{17/20} & [64.0,94.8] & \textbf{88.2} \\
          & Clean-only & 30000 & -- & 19/20 & [76.4,99.1] & 100.0 \\
        \midrule
        \multirow{5}{*}{ACT/C}
          & $\thetaInit$ & full & -- & 5/20 & [11.2,46.9] & 0.0 \\
          & Supp. & 200 & 4800 & 6/20 & [14.5,51.9] & 6.7 \\
          & Clean FT & 200 & 3200 & 9/20 & [25.8,65.8] & 26.7 \\
          & \method & 200 & 3200 & \textbf{18/20} & [69.9,97.2] & \textbf{86.7} \\
          & Clean-only & 50000 & -- & 20/20 & [83.9,100.0] & 100.0 \\
        \midrule
        \multirow{4}{*}{$\pi_{0.5}$/T}
          & $\thetaInit$ & full & -- & 2/20 & [2.8,30.1] & 0.0 \\
          & Clean FT & 250 & 8000 & 5/20 & [11.2,46.9] & 21.4 \\
          & \method & 250 & 8000 & \textbf{15/20} & [53.1,88.8] & \textbf{92.9} \\
          & Clean-only & 30000 & -- & 16/20 & [58.4,91.9] & 100.0 \\
        \bottomrule
    \end{tabular}
\end{table}

\begin{table}[t]
    \caption{Heterogeneous-Defect30 StackCube (\%, mean $\pm$ SD over three seeds).}
    \label{tab:heterogeneous}
    \centering
    \scriptsize
    \setlength{\tabcolsep}{3.0pt}
    \begin{tabular}{lrrrr}
        \toprule
        Method & Success$\uparrow$ & GapRec$\uparrow$ & RR$\uparrow$ & RG$\downarrow$ \\
        \midrule
        Clean-only & $89.3{\pm}1.5$ & 100.0 & -- & -- \\
        $\thetaInit$ & $63.3{\pm}3.5$ & 0.0 & -- & -- \\
        FT-CM & $66.0{\pm}5.6$ & $11.3{\pm}16.3$ & $40.8{\pm}4.8$ & $19.6{\pm}9.9$ \\
        Suppression & $64.0{\pm}15.6$ & $5.7{\pm}46.8$ & $53.2{\pm}7.8$ & $30.1{\pm}20.2$ \\
        \method & $\mathbf{81.0{\pm}1.7}$ & $\mathbf{68.6{\pm}7.7}$ &
        $\mathbf{70.0{\pm}0.3}$ & $\mathbf{12.7{\pm}2.0}$ \\
        \bottomrule
    \end{tabular}
\end{table}

\subsection{Real-Robot Repair Performance}

\begin{figure*}[t]
    \centering
    \includegraphics[width=0.98\textwidth]{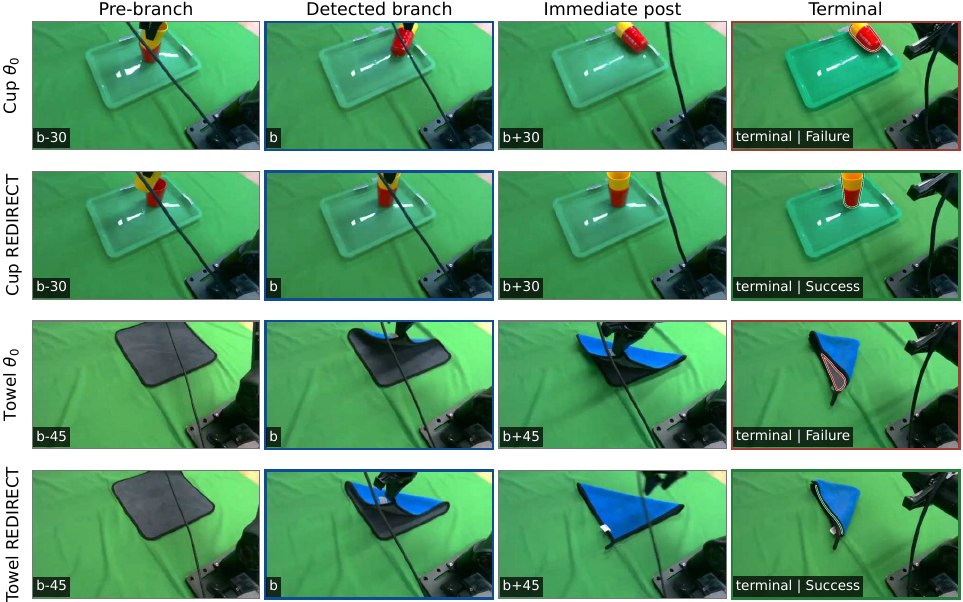}
    \caption{Representative real-robot rollouts aligned to the frozen manifest branch for the corresponding placement. Columns show pre-branch, branch, immediate post-branch, and terminal observations; rows compare $\thetaInit$ and \method on cup insertion and towel folding. Blue, red, and green borders mark the aligned branch, labeled failure, and labeled success, respectively. Table~\ref{tab:realrobot} reports all trials.}
    \label{fig:realrobot}
\end{figure*}

\method achieves 17/20, 18/20, and 15/20 successes on $\pi_{0.5}$ cup insertion, ACT cup insertion, and $\pi_{0.5}$ towel folding, compared with 2/20, 9/20, and 5/20 for matched-budget FT (Table~\ref{tab:realrobot}). Comparisons with FT yield Fisher $p=3.36\times10^{-6}$ and $p=0.0057$ for the cup settings, and exact McNemar $p=0.0129$ for towel folding. Figure~\ref{fig:realrobot} illustrates representative changes in release and placement behavior.

Increasing $\pi_{0.5}$ cup FT to four times the budget raises success to 15/20. Thus, clean-data FT can recover performance with more optimization, while \method achieves substantial recovery within the smaller budget.

\subsection{Computational Cost}

\method uses 1.0\%, 0.8\%, and 1.67\% of full-training SB in simulation, physical ACT, and physical $\pi_{0.5}$, respectively. Its physical optimization time is 1.29\% of full-retraining time for ACT and 1.55--1.57\% for $\pi_{0.5}$ (Table~\ref{tab:wallclock}). These timings cover optimization only.

\subsection{Branch Localization and Corrective Targets}

The estimated onset has a mean absolute error of 8.38 frames and a median of 0, with 74.4\% of episodes within $\pm5$ frames (Fig.~\ref{fig:localization_compute}(b)).

Random Window changes only the edit center and reduces success (Table~\ref{tab:target_ablations}). Local Clean FT reaches 77.8\% despite using retrieved clean pairs: it trains at clean observations, while \method provides corrective action targets at the original problematic observations. This comparison supports directly supervising the observations associated with the failure, which retained-data FT does not explicitly target.

All target-construction alternatives in Table~\ref{tab:target_ablations} yield lower success and higher regression, supporting a fixed source trajectory with consistent relative timing.

\begin{figure*}[t]
    \centering
    \includegraphics[width=0.96\textwidth]{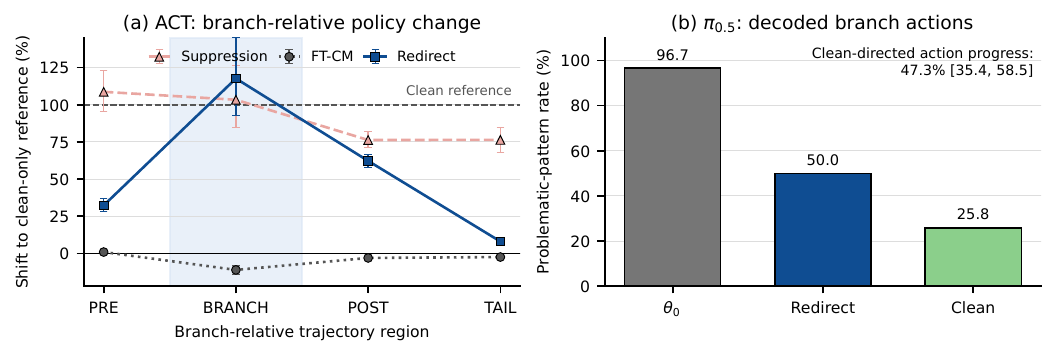}
    \caption{Branch-local policy diagnostics on problematic real trajectories.  (a) ACT branch-relative original-label loss shift from $\thetaInit$ (0\%) toward clean-only (100\%); shading marks BRANCH observations.  (b) $\pi_{0.5}$ problematic-pattern assignment and clean-directed action progress. Error bars in (a) and brackets in (b) are 95\% episode-bootstrap intervals.}
    \label{fig:mechanism}
\end{figure*}

\subsection{Behavioral Preservation}

Adding retained replay to Edit Only raises success from 39.0\% to 86.0\%. Full \method achieves the highest success and lowest regression (Table~\ref{tab:core_ablations}).

On 90 held-out clean twins, \method reduces action MAE compared with Edit Only, from 0.233 to 0.138 on PickCube, 0.222 to 0.120 on PushCube, and 0.188 to 0.159 on StackCube.

Figure~\ref{fig:mechanism} shows that ACT original-label loss changes concentrate around BRANCH and POST, while $\pi_{0.5}$ reduces problematic-pattern assignment and moves decoded actions toward the clean-only reference. In the ACT deployment audit, BRANCH/TAIL MAE relative to $\thetaInit$ is 7.44/3.05\% for \method, 1.36/1.42\% for FT, and 4.93/11.16\% for suppression. \method reduces POST distance to clean-only from 16.59\% to 8.26\%, while retained-label MAE increases from 2.38\% to 2.84\%.

Projection activates in 24.1\% of updates; exact AdamW replay finds retain conflict in 9 of 2,250 steps (0.4\%). Preservation is therefore supported empirically, without a guarantee under AdamW.

\begin{table}[!t]
    \caption{Optimization wall-clock and percentage of clean-only full retraining; ranges span tasks.}
    \label{tab:wallclock}
    \centering
    \scriptsize
    \setlength{\tabcolsep}{2.0pt}
    \begin{tabular}{llrrrr}
        \toprule
        Policy & Method & Upd. & SB & Opt. (s) & \%Full \\
        \midrule
        \multirow{4}{*}{$\pi_{0.5}$}
          & Full train & 30000 & 480000 & 55807--57276 & 100.0 \\
          & Clean FT & 250 & 8000 & 832.8--833.6 & 1.45--1.49 \\
          & Clean FT$^{*}$ & 1000 & 32000 & 3318.2 & 5.95 \\
          & \method & 250 & 8000 & 878.0--890.6 & 1.55--1.57 \\
        \midrule
        \multirow{4}{*}{ACT}
          & Full train & 50000 & 400000 & 13645.0 & 100.0 \\
          & Clean FT$^{\dagger}$ & 200 & 3200 & 192.5 & 1.41 \\
          & Supp.$^{\dagger}$ & 200 & 4800 & 182.1 & 1.33 \\
          & \method & 200 & 3200 & 176.3 & 1.29 \\
        \bottomrule
    \end{tabular}
    \vspace{1pt}
    \parbox{0.96\columnwidth}{\scriptsize $^{*}$1000-update FT is cup only. $^{\dagger}$Times are bounded by the first/last timestamped checkpoint artifacts because the optimizer report was not archived.}
\end{table}

%% file: sections/06_discussion.tex
\section{Discussion}
\label{sec:discussion}

\paragraph{Why a branch exists}
For the local defects studied here, branch localization separates the problematic continuation from shared behavior, allowing repair below the episode level despite unpaired trajectory variation.

\paragraph{Why 1\% is enough}
\method provides corrective targets at problematic observations, which clean-data fine-tuning does not directly supervise. The Local Clean FT comparison supports this targeted use of the update budget (Table~\ref{tab:target_ablations}).

\paragraph{A fix needs an executable normal future}
Suppression lacks a replacement target. The single-source medoid supplies a recorded continuation rather than splicing separately retrieved actions or averaging across modes.

\paragraph{Scope}
Evidence covers local, in-distribution differences within one task, object semantics, controller, and action interface. One invocation repairs the dominant branch; repairing several separated mistakes remains future work. AdamW preservation is empirical, supported by paired regressions, held-out clean twins, temporal diagnostics, and exact step audits.

%% file: sections/07_conclusion.tex
\section{Conclusion}
\label{sec:conclusion}

\method localizes behavioral branches, retrieves coherent replacement continuations, and anchors updates on retained demonstrations. With 1\% of full-training sample-backward compute, it raises the simulation success rate from 68.2\% to 90.3\%, recovers 88.8\% of the clean gap, and transfers across ACT, $\pi_{0.5}$, and two real-robot tasks. In large datasets, where problematic subsets are difficult to identify, low-cost repair could support iterative testing of candidate subsets. Coupling subset selection with repeated repair and evaluation may help guide data refinement without repeated full retraining.